\documentclass{IEEEcsmag}

\usepackage[colorlinks,urlcolor=blue,linkcolor=blue,citecolor=blue]{hyperref}

\usepackage{upmath}

\jvol{XX}
\jnum{XX}
\paper{8}
\jmonth{May}
\jname{}
\pubyear{2020}

\begin{document}

\title{Cyberbullying Detection with Fairness Constraints}

\author{Oguzhan Gencoglu}
\affil{Tampere University, Faculty of Medicine and Health Technology, Tampere, Finland \\ \texttt{oguzhan.gencoglu@tuni.fi}}

\begin{abstract}
Cyberbullying is a widespread adverse phenomenon among online social interactions in today's digital society. While numerous computational studies focus on enhancing the cyberbullying detection performance of machine learning algorithms, proposed models tend to carry and reinforce unintended social biases. In this study, we try to answer the research question of ``Can we mitigate the unintended bias of cyberbullying detection models by guiding the model training with fairness constraints?''. For this purpose, we propose a model training scheme that can employ fairness constraints and validate our approach with different datasets. We demonstrate that various types of unintended biases can be successfully mitigated without impairing the model quality. We believe our work contributes to the pursuit of unbiased, transparent, and ethical machine learning solutions for cyber-social health.
\end{abstract}

\maketitle

\chapterinitial{Cyberbullying} (CB) has emerged as a serious and widespread adverse phenomenon among online social interactions in today's digital society. Estimates indicate that between 15-35\% of young people have been victims of cyberbullying and between 10-20\% of individuals admit to having cyberbullied others \cite{hinduja2010bullying}. Cyberbullying can be observed in the form of aggression, harassment, derogation, dehumanization, toxicity, profanity, racism, sexism, call for violence, hate speech, threatening, or any combination thereof. Consequences of cyberbullying are severe as victims were shown to be at a greater risk of depression, anxiety, negative emotions, self-harm, suicidal behaviors, and psychosomatic symptoms than non-victims. %To a lesser extent, perpetrators of cyberbullying are at risk of suicidal behaviors and suicidal ideation when compared with nonperpetrators \cite{john2018self}.
% Digital harm directed to oneself is not uncommon either. In \cite{patchin2017digital}, about 6\% of adolescents were shown to display digital self-harm by anonymously posting something online about themselves that was mean.

Cyberbullying detection, i.e., automatically classifying online text messages based on bullying content with machine learning, has been proposed by numerous studies to detect and prevent this phenomenon \cite{rosa2019automatic}. Accurate and timely detection of cyberbullying remains an unsolved task as cyberbullying is immensely context-dependent and occasionally subtle (e.g. hidden mockery and sarcasm). In addition, significant amount of cyberbullying takes place in private conversations without any publicity. Statistical drifts and overall changes in popular culture, politics, human language, and media of communication further increase the complexity of this task. %Consequently, preventing cyberbullying still calls for better tools including algorithmic ones. 

While cyberbullying detection research focus predominantly on increasing the overall accuracy and timeliness of detection, recent studies reveal that proposed systems carry significant amount of unintended bias against certain groups of users, demographics, languages, dialects, terms, phrases, or topics. For instance, terms like ``gay'' and ``jew'', present in an informative or conversational context, tend to incline the model predictions towards ``hate speech'' or ``high toxicity''. Similarly, tweets in African-American English were shown to be more likely to be classified as abusive or offensive compared to other tweets \cite{davidson2019racial}. As such groups can be minorities among the whole population, trying to maximize the detection accuracy and evaluating approaches solely on overall detection performance exacerbates the bias even further. The source of these undesirable biases can be machine learning algorithms, datasets used for training, pre-trained language models, human annotators, or typically a combination thereof.

In this study, we try to answer the following two research questions: ``\textit{Can we mitigate the unintended bias of cyberbullying detection models by guiding the model training with fairness constraints}''? and ``\textit{If so, how much does such bias mitigation impair the cyberbullying detection performance}''?. We hypothesize that, if formulated in a fairness context, recent advancements in constrained optimization \cite{cotter2019two} can be successfully utilized to mitigate the unintended bias in cyberbullying detection models without hindering overall detection performance. To test our hypothesis, we formulate frequently used fairness assessment metrics from the literature as constraints that can be utilized during model training. We then validate the proposed approach with 4 different datasets and varying bias contexts such as gender bias, language bias, and recency bias. We show that our approach enables the discovery of models with more equitable performance across different groups of interest while improving overall and group-specific cyberbullying detection performances. In addition, our approach is agnostic to data modality (e.g. text, image, graph, tabular) and does not require group labels during inference. We release the full source code of our study along with the trained models (\url{https://github.com/ogencoglu/fair_cyberbullying_detection}). We believe our work contributes to the pursuit of unbiased, transparent, ethical, and accountable machine learning solutions for cyber-social health.

\section{THE PURSUIT OF FAIRNESS}

Previous studies of cyberbullying detection utilized numerous machine learning approaches varying from rule-based systems to deep learning \cite{rosa2019automatic}. Reported performances of the proposed approaches differ significantly between studies as well ($F_1$ scores in the range of 0.4-0.8 for most studies), depending on the experimentation dataset and definition of CB \cite{rosa2019automatic}. As in several other natural language processing (NLP) tasks, most recent CB detection studies utilize deep neural networks and transfer learning, i.e., employing a pre-trained language model to retrieve informative numerical representations of the textual data (e.g. representations of tokens, words, sentences, or paragraphs) and training a classifier with these representations \cite{nozza2019unintended,huang2020multilingual}.

Computational methods can propagate and even reinforce social biases. A machine learning model is considered to contain unintended bias if it performs better for some demographic groups or groups of interest than others. In natural language processing, unintended biases emerge from bias in datasets, bias in distributed word embeddings (e.g. word2vec, GloVe), bias in contextual word embeddings (e.g. BERT, ELMo, GPT-2), bias in sentence embeddings, bias in machine learning algorithms, or bias in human annotators. For instance, a recent study by Nadeem et al. show that stereotypical bias is amplified as pretrained language models (e.g. BERT, RoBERTa, XLNet, GPT-2) become stronger \cite{nadeem2020stereoset}. While there have been several studies of CB detection and several studies of algorithmic bias analysis in machine learning, only a handful of studies attempt to combine the two. To the best of our knowledge, an extensive systematic review of bias quantification and mitigation methods in cyberbullying detection does not exist as of August 2020.

In cyberbullying detection context, even though studies show that CB is more common to be perpetrated by certain identity groups (e.g. young males) and perpetrated against certain identity groups (e.g. non-heterosexuals), detection performance of an unbiased model in terms of \textit{equality of odds} is expected to be the same among all groups as well as with respect to the overall population. For instance, an unbiased model should not classify a text containing \textit{identity terms} (e.g. black, jew, gay, white, christian, transgender) in an innocuous context as ``cyberbullying'' only because such terms appear frequently in cyberbullying incidents. Nonetheless, previous studies show that models tend to label non-toxic statements containing identity terms as toxic, manifesting a false positive bias \cite{dixon2018measuring,park2018reducing}. Bias is observed against different languages and dialects as well. Classifiers were shown to predict African-American English tweets more likely to be abusive or offensive than Standard American English \cite{davidson2019racial}.

Proposed approaches for mitigation of unintended bias in CB detection predominantly involve \textit{balancing} the training data in some statistical sense, such as oversampling, data augmentation, or sample weighting. For instance, Dixon et al. add additional non-toxic examples containing identity terms from Wikipedia articles to training data \cite{dixon2018measuring}. Badjatiya et al. detect bias-sensitive words and replace them with neutral words or tokens to reduce stereotypical bias for hate speech detection \cite{badjatiya2019stereotypical}. Nozza et al. sample additional data from an external corpus to augment training data for misogyny detection \cite{nozza2019unintended}. Similarly, Park et al. augment the training data by identifying male entities and swapping them with equivalent female entities and vice-versa in order to reduce gender bias in abusive language detection \cite{park2018reducing}. These methods, inevitably, introduce additional calibration hyper-parameters that are highly influential both on the performance of cyberbullying detection and bias mitigation. For instance, increasing the prevalence of disadvantaged groups during training may impair overall classification performance due to high distortion of decision boundary in the feature space. Furthermore, as the statistical properties of the modeled phenomena change over time in unforeseen ways (also known as \textit{concept drift}), proposed calibration parameters become obsolete with time in real-life applications.

%Several statistically significant correlates of involvement in digital self-harm were identified including sexual orientation \cite{patchin2017digital}.

In this study, we formulate the task of training more equitable CB detection models as a constrained optimization problem. Just like standard deep neural network training, the optimization task corresponds to adjusting model weights by minimizing a loss function, iterated over the training data. However, by imposing fairness constraints during model training, we enforce training to converge to a more equitable solution for the chosen groups of interest. Essentially, our method does not alter the training data and can be used simultaneously with the abovementioned data debiasing methods for further bias mitigation.

%To mitigate gender bias in computer vision tasks, injects constraints on the output distribution during inference to ensure the model predictions follow the distribution observed from the training data (requiring a joint inference over all test data) \cite{zhao2017men}. The outputs of the classifier are then calibrated to enforce fairness. By decoupling the training from the fairness enforcement, this procedure may not lead to the best trade-off between fairness and accuracy.

\section{METHODS}
\subsection{Datasets}

We validate our methods by performing experiments with 4 publicly available datasets collected from 4 different media. In order to further test the generalization power of the proposed bias mitigation approach, we appoint different identity group contexts for each experiment including identity-related groups, language-related groups, and date-related groups. Datasets differ from each other regarding total number of samples, overall proportion of cyberbullying samples, and group-specific proportions of cyberbullying samples.

Focus of the first experiment is \textit{gender bias}. We demonstrate our method on the Perspective API's Jigsaw dataset of 403,957 comments with toxicity and identity annotations \cite{borkan2019nuanced}. We select \textit{male} and \textit{female} identities as groups of interest. Percentage of the male and female groups are 11.0\% and 13.2\%, respectively. Toxicity proportion among the male and female groups are 15.0\% and 13.7\%, respectively and overall toxicity proportion in the dataset is 11.4\%. %We treated a given comment as cyberbullying if the mean score of annotations is above 0.5 (minimum and maximum being 0 and 1, respectively).

Focus of the second experiment is \textit{language bias}. We employ a multilingual dataset of 107,441 tweets with hate speech annotations in 5 languages, namely \textit{English}, \textit{Italian}, \textit{Polish}, \textit{Portuguese}, and \textit{Spanish} \cite{huang2020multilingual}. Percentage of English, Italian, Polish, Portuguese, and Spanish tweets are 78.1\%, 5.4\%, 10.2\%, 1.8\%, and 4.6\%, respectively. Hate speech proportion among English, Italian, Polish, Portuguese, and Spanish tweets are 26.2\%, 19.5\%, 9.0\%, 20.3\%, and 39.7\% respectively and overall hate speech proportion in the dataset is 24.6\%.

Third experiment has a more practical focus of \textit{date bias} or \textit{recency bias}. As popular culture and languages change, new phrases of insult may emerge, leading to underperforming models on more recent observations. We would like to first demonstrate, then mitigate the bias of recency in CB classifiers. The dataset utilized for validating our mitigation approach is WikiDetox dataset (shortly referred as Wiki from now on) from Wikipedia Talk Pages \cite{wulczyn2017ex}. This dataset consists of three distinct corpora of comments from English Wikipedia Talk pages posted between 2001-2016 and labeled for aggression, personal attack, and toxicity. We select 77,972 comments that have all of the annotations regarding aggression, personal attack, and toxicity for this study. The groups of interest are comments posted between the years \textit{2001-2014} and \textit{2015-2016} (\textit{recent} group). Percentage of the 2001-2014 and 2015-2016 groups are 93.8\% and 6.2\%, respectively. CB proportion among the 2001-2014 and 2015-2016 groups are 20.9\% and 19.3\%, respectively and overall CB proportion in the dataset is 20.8\%.

Final experiment concentrates on bias among hate speech on the basis of \textit{religion}, \textit{race}, and \textit{nationality}. For this purpose, we utilize the largest corpus of hate speech to date with theoretically-justified annotations called Gab Hate Corpus \cite{kennedy2020gab}. Dataset consists of 27,665 posts from the social network platform Gab. Percentage of hate speech on the basis of religion, race, and nationality are 5.1\%, 6.5\%, and 4.8\%, respectively. Hate speech proportion among the religion, race, and nationality groups are 51.2\%, 53.7\%, and 39.6\%, respectively and overall hate speech proportion in the dataset is 8.0\%.

\subsection{Cyberbullying Detection Model}

We utilize recently introduced deep neural network architectures from NLP research, i.e. transformer networks, for performing binary classification of cyberbullying (cyberbullying or not). We employ a multilingual language model, sentence-DistilBERT \cite{reimers2019sentence}, for extracting \textit{sentence embeddings} to represent each post/comment. We then train a simple 3-layer (2 dense layers of size 512 and 32, respectively and an output layer) fully-connected neural network using the sentence embeddings as input features. % We train the baseline (unconstrained) and fairness-constrained model

We chose sentence embeddings instead of fine-tuning on token embeddings (e.g. out of original BERT model or its variants) firstly due to its low computational requirements \cite{reimers2019sentence}. Secondly, even though it is possible to represent sentences with a standard BERT model as well (e.g. by averaging the token embeddings out of the output layer), for several semantic textual similarity tasks BERT sentence embeddings were shown to be inferior to embeddings specifically trained for representing sentences \cite{reimers2019sentence}. We chose a multilingual model to be able to extract useful representations from the Twitter dataset as well, as it includes tweets from 5 different languages.

% Since we are solving a constrained optimization problem, the final iterate from training may not always be the best solution, and an intermediate iterate may do a better job of optimizing objective while satisfying the constraints.

% if the groups we wish to constraint are all small in size, and its likely that the individual minibatches contains very few examples from each group. Hence the gradients we compute during training will be noisy, and result in the model converging very slowly.

For each dataset, we train baseline (unconstrained) and fairness-constrained models in a mini-batch manner for 75 epochs with a batch size of 128 with Adam optimizer (learning rate of $5 \times 10^{-4}$). In order to avoid overfitting, models at the epoch that maximizes the $F_1$ score on the validation set (70\%-15\%-15\% training-validation-test split) is selected as final model for each training. Baseline and fairness-constrained models are trained, validated, and tested with the same data and employ the same neural network architecture with the same hyper-parameters in order to establish fair comparison. We would like to emphasize that instead of maximizing the cyberbullying detection performance, mitigation of unintended bias while maintaining the detection performance has been the main focus of this study. Therefore, we neither experimented with different text preprocessing schemes, pretrained models, sophisticated neural network architectures, nor performed any hyper-parameter tuning.

%Essentially, baseline models try to minimize binary cross-entropy (chosen loss function) directly and maximize $F_1$ score indirectly while constrained models try to maximize $F_1$ score directly. 

\subsection{Fairness Constraints}

Fairness of machine learning models can be assessed by the metrics \textit{False Negative Equality Difference} (FNED) and \textit{False Positive Equality Difference} (FPED), as done in several studies \cite{huang2020multilingual,dixon2018measuring,park2018reducing}. FNED/FPED is simply the sum of deviations of group-specific False Negative Rates (FNRs)/False Positive Rates (FPRs) from the overall FNR/FPR. For $N$ groups and set of observations belonging to each group being $G_{i \; \in \; \{1, \cdots, N\}}$,
\begin{equation}
\begin{aligned}
FNED = \sum_{i \in \{1,\cdots,N\}}{\left|FNR - FNR_{G_i}\right|} \\
FPED = \sum_{i \in \{1,\cdots,N\}}{\left|FPR - FPR_{G_i}\right|}.
\end{aligned}
\label{eq1}
\end{equation}
In essence, group-specific error rates of fairer models deviate less from the overall error rates and consequently from each other, approaching an ideal equality of odds. Being related to the \textit{equalized odds} fairness criteria, sum of FNED and FPED corresponds to the total unintended bias of a model (ideally zero).

Neural network training can be formulated as finding a set of parameters (network weights), $\theta$, minimizing an objective function, $f_{L}$:
\begin{equation}
\qquad\qquad\qquad \min_{\theta} f_{L}(\theta),
\label{eq2}
\end{equation}
where $f_{L}$ is typically binary cross-entropy loss for binary classification tasks (such as here) and minimization is performed by backpropagation algorithm on the training set. In fairness-constrained neural network training, we would like to minimize the same function in Equation (2) while constraining the deviation of each group-specific FNR/FPR from the overall FNR/FPR in order to decrease the unintended bias, i.e.,
\begin{equation}
\begin{aligned}
\min_{\theta} f_{L}(\theta) \qquad &\\
\mbox{subject to} \;\;\;
\left|FNR - FNR_{G_1}\right|&<\tau_{FNR} \\
\left|FPR - FPR_{G_1}\right|&<\tau_{FPR} \\
\cdots \;\;\;\;\;\; \\
\left|FNR - FNR_{G_N}\right|&<\tau_{FNR} \\
\left|FPR - FPR_{G_N}\right|&<\tau_{FPR},
\end{aligned}
\label{eq3}
\end{equation}
where $\tau_{FNR}$ and $\tau_{FPR}$ are allowed deviations (corresponding to biases) from overall FNRs and FPRs, respectively. 

In principle, implementation of Equation (3) involves $2N$ constraints (one for FNR and one for FPR for each group of interest). As constraints are inherently guiding the neural network training in our method, convergence becomes more difficult with increasing number of constraints. Therefore, it is beneficial to express the same inequalities with fewer number of constraints such as:
\begin{equation}
\begin{aligned}
\min_{\theta} f_{L}(\theta) \qquad &\\
\mbox{s.t.}  \;\;\; \max \{ FNR_{G_1}, \cdots\}-FNR&<\tau_{FNR} \\
FNR-\min\{ FNR_{G_1}, \cdots\}&<\tau_{FNR}\\
\max \{ FPR_{G_1}, \cdots\}-FPR&<\tau_{FPR} \\
FPR-\min\{ FPR_{G_1}, \cdots\}&<\tau_{FPR}.
\end{aligned}
\label{eq4}
\end{equation}
By constraining the upper and lower bounds of group-specific FNRs/FPRs with respect to overall FNR/FPR, number of constraints can be decreased strictly to four. Proposed constraints in Equation (4) correspond to direct incorporation of fairness metrics into neural network training in Equation (1). Essentially, such worst-case oriented reformulation of the fairness constraints can be considered as a \textit{robust optimization} problem.

Typical approaches that perform constrained optimization are based on \textit{Lagrange multipliers} where a saddle point to the Lagrangian will correspond to an optimal solution (under fairly general conditions). However, this no longer holds for non-convex settings such as neural network training with gradient descent. In fact, gradient descent may not converge at all and oscillate in such settings. Furthermore, constrained neural network training (as well as training of several other machine learning algorithms) requires differentiable constraints and loss functions in order to compute the gradients for minimizing a given empirical loss. Considering fairness constraints are data-dependent \textit{counts} and \textit{proportions} (e.g. false positive rate), gradients (or subgradients) are either unavailable or always zero. To overcome this challenge, we propose utilizing recent advancements in solving constrained, non-convex optimization problems in which constraints can be non-differentiable (as in our study) \cite{cotter2019two}. Cotter et al. formulate such a problem as a two-player zero-sum game and introduce \textit{proxy constraints} that are differentiable approximations of the original constraints \cite{cotter2019two}. During training, each proxy constraint function is penalized in such a proportion that the penalty satisfies the original (exact) constraint.

In practice, implementation of constrained training is performed via two data streams. First data stream feeds random data batches and corresponding binary labels (cyberbullying or not) to the model. Second data stream traces the samples belonging to the groups of interest in a given batch and computes the constraint violations, which in return will be used for additional penalty on top of the misclassification penalization from the first stream. For instance, if the interest groups are \textit{race}, \textit{religion}, and \textit{nationality} and $\tau_{FNR}$ is set to 0.1, the model will be penalized if the maximum of $(FNR_{race}, FNR_{religion}, FNR_{nationality})$ goes above $FNR_{overall} + 0.1$ or if the minimum of those goes below $FNR_{overall} - 0.1$ during each iteration. Same logic applies to FPRs. Once the model is trained, access to group attributes is not needed during inference, i.e., constrained models can be used in the same manner as unconstrained models. We implemented our experiments using TensorFlow framework in Python 3.7.

We set $\tau_{FNR}$ to 0.02, 0.15, 0.005, and 0.1; $\tau_{FPR}$ to 0.03, 0.1, 0.005, and 0.15 for Jigsaw, Twitter, Wiki, and Gab experiments, respectively. Even though a hypothetical, perfectly unbiased model would have zero deviation between group-specific rates and overall rates, it is important to remember that perfect \textit{calibration} (outcomes being independent of group attributes) can not be satisfied simultaneously with the balance for false negative and false positive rates. Therefore, forcing the model to satisfy overly-conservative fairness constraints (very small $\tau_{FNR}$ and $\tau_{FPR}$) or multiple fairness notions simultaneously may not only impair classification performance, but may also result in training failing to converge (oscillating loss curves). Furthermore, overconstraining the model may result in convergence to a trivial unbiased solution, such as a model outputting the same prediction regardless of the input.

\subsection{Evaluation}

\begin{figure}
\centerline{\includegraphics[clip,trim={0.1cm 4cm 12.5cm 0.1cm},width=18.5pc]{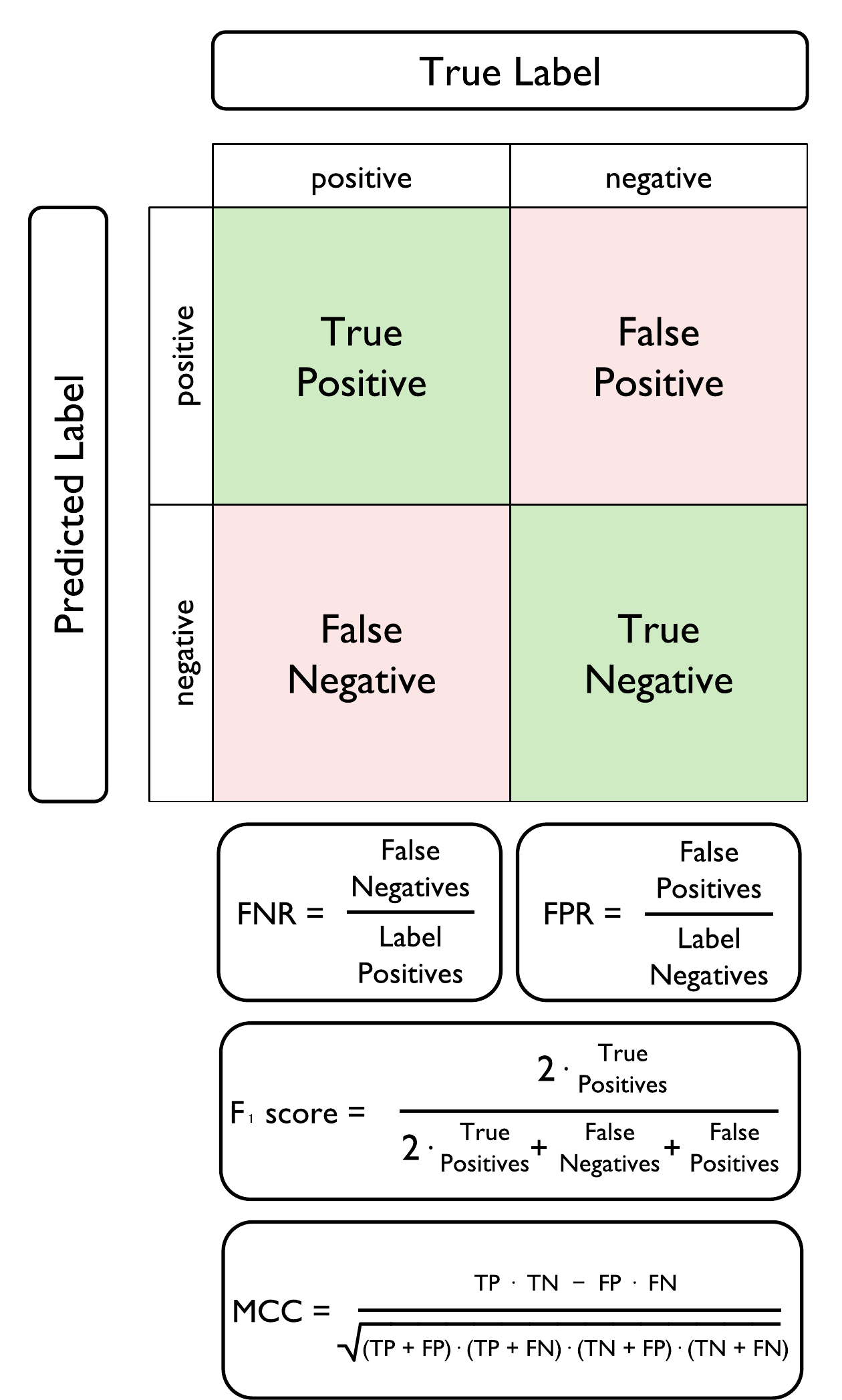}}
\caption{Definition of common performance metrics for cyberbullying detection.}
\end{figure}

For evaluation of our methods we randomly split every dataset into training, validation, and test sets with 70\%, 15\%, and 15\% splits, respectively (random seed is fixed for every experiment for reproducibility and fair comparison). All results reported in this study are calculated from the test set. Due to its popularity, we report the standard metric of $F_1$ score of our binary classifiers. As accuracy, area under the receiver operating characteristic curve, and $F_1$ score may lead to misleading conclusions on imbalanced datasets (such as here), we evaluate our models with Matthews correlation coefficient (MCC) which does not exhibit such drawback \cite{chicco2020advantages}. MCC generates a high score only if the binary predictor was able to correctly predict the majority of positive instances and the majority of negative instances \cite{chicco2020advantages}. We use MCC for comparing group-specific detection performances between the baseline and fairness-constrained models as well. Furthermore, we perform McNemar's test (also known as \textit{within-subjects chi-squared test}) to test whether a statistically significant difference between baseline (unconstrained) and constrained models exists in terms of cyberbullying classification accuracy. Definitions of False Negative Rate, False Positive Rate, $F_1$ score, and Matthews correlation coefficient is depicted in \textbf{Figure 1}. For quantifying the unintended bias of our models, we use the sum of FNED and FPED, calculated on the test set.

\begin{figure*}
\centering{\includegraphics[clip,trim={0cm 0cm 0cm 0cm},width=26.6pc]{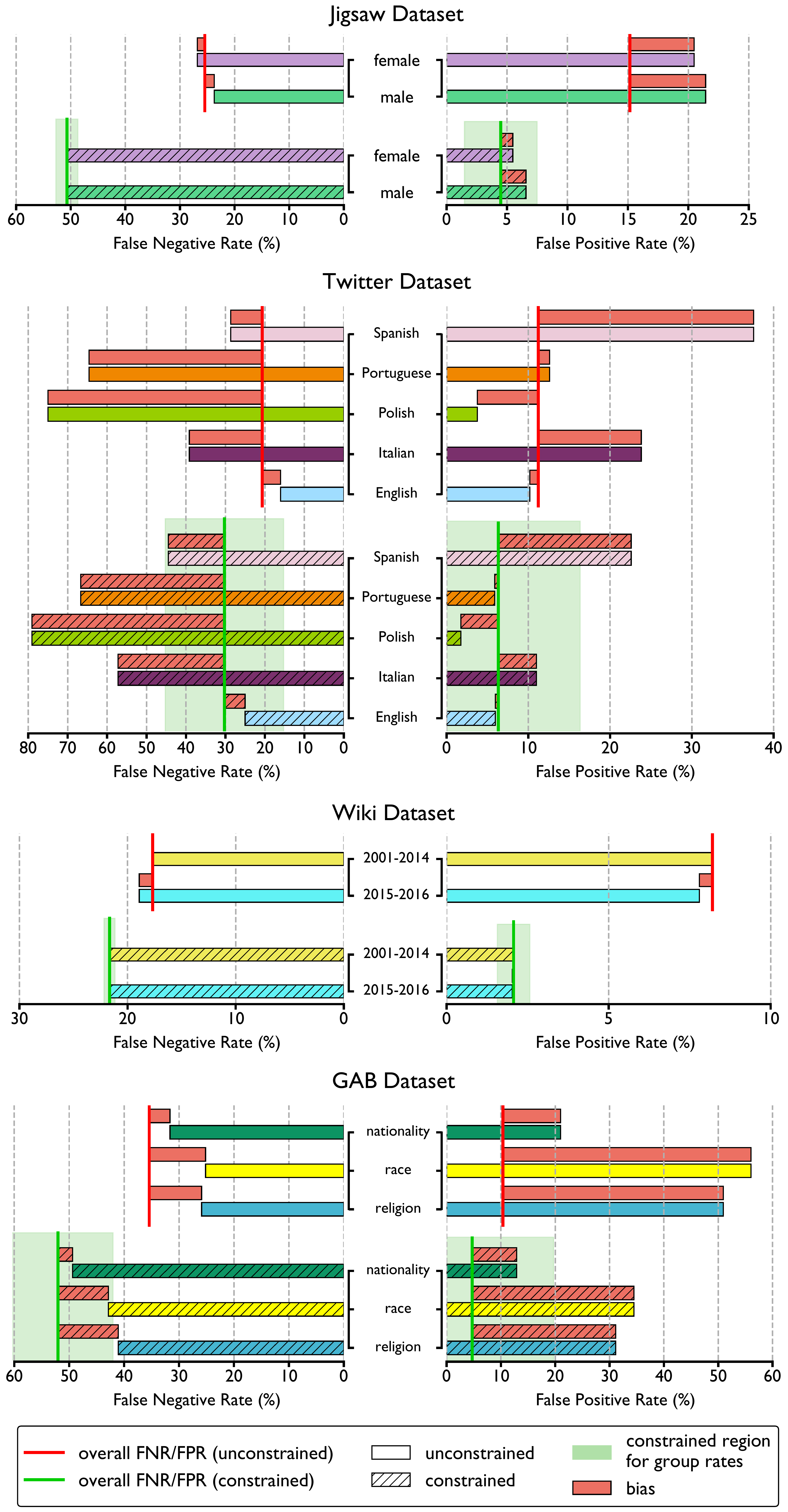}}
\caption{False negative rates, false positive rates and biases for each group for unconstrained and fairness-constrained cyberbullying detection models on the test set among 4 experiments.}
\end{figure*}

\section{RESULTS}

\begin{table*}
\caption{Performance of baseline (unconstrained) and fairness-constrained cyberbullying detection models evaluated on the test set and corresponding biases (P = Precision, R = Recall).}
\small
\begin{tabular*}{35pc}{|p{24.4pt}|p{9pt}|p{13pt}|p{9pt}|p{9pt}|p{16pt}|p{16pt}|p{16pt}||p{9pt}|p{13pt}|p{9pt}|p{9pt}|p{16pt}|p{16pt}|p{16pt}|p{31.9pt}|}
\cline{2-15}
\multicolumn{1}{c|}{} & \multicolumn{7}{c||}{Baseline (unconstrained) Model} & \multicolumn{7}{c|}{Fairness-constrained Model}\\
\cline{1-16}
\vspace{0pt} Dataset & \vspace{0pt} $F_1$ & \vspace{0pt} \hspace*{-1.4mm} MCC & \vspace{0pt} \centering{P} & \vspace{0pt} \centering{R} & \vspace{0pt} \hspace*{-1.5mm} FNED & \vspace{0pt} \hspace*{-1.3mm} FPED & \centering{Total Bias} & \vspace{0pt} $F_1$ & \vspace{0pt}  \hspace*{-1.4mm} MCC & \vspace{0pt} \centering{P} & \vspace{0pt} \centering{R} & \vspace{0pt} \hspace*{-1.5mm} FNED & \vspace{0pt} \hspace*{-1.3mm} FPED & \centering{Total Bias} & \hfil \footnotesize{Bias} \newline \footnotesize{Decr. (\%)}\\
\cline{1-16}
\hfil Jigsaw  & \hspace*{-1.3mm} 0.51 & 0.46 & \hspace*{-1.4mm} 0.39 & \hspace*{-1.4mm} 0.75 & 0.031 & 0.116 & 0.147 & \hspace*{-1.3mm} 0.54 & 0.49 & \hspace*{-1.4mm} 0.59 & \hspace*{-1.4mm} 0.49 & 0.001 & 0.031 & 0.031 & \hfil 78.7\\
\hfil Twitter & \hspace*{-1.3mm} 0.74 & 0.65 & \hspace*{-1.4mm} 0.70 & \hspace*{-1.4mm} 0.79 & 1.293 & 0.488 & 1.781 & \hspace*{-1.3mm} 0.74 & 0.66 & \hspace*{-1.4mm} 0.78 & \hspace*{-1.4mm} 0.70 & 1.316 & 0.263 & 1.579 & \hfil 11.3\\
\hfil Wiki    & \hspace*{-1.3mm} 0.77 & 0.71 & \hspace*{-1.4mm} 0.73 & \hspace*{-1.4mm} 0.82 & 0.013 & 0.004 & 0.018 & \hspace*{-1.3mm} 0.84 & 0.81 & \hspace*{-1.4mm} 0.91 & \hspace*{-1.4mm} 0.78 & 0.001 & 0.000 & 0.001 & \hfil 94.9\\
\hfil Gab     & \hspace*{-1.3mm} 0.45 & 0.41 & \hspace*{-1.4mm} 0.34 & \hspace*{-1.4mm} 0.65 & 0.236 & 0.968 & 1.203 & \hspace*{-1.3mm} 0.47 & 0.42 & \hspace*{-1.4mm} 0.46 & \hspace*{-1.4mm} 0.48 & 0.228 & 0.644 & 0.872 & \hfil 27.5\\
\cline{1-16}
\end{tabular*}
\label{table1}
\end{table*}

\begin{table*}
\caption{Group-specific Matthews correlation coefficients for baseline (unconstrained) and fairness-constrained models (better performance highlighted).}
\hspace{0.1cm}
\small
\begin{tabular*}{4.6pc}{|>{\centering}m{38pt}|c|c|}
\cline{2-3}
\multicolumn{1}{>{\centering}m{25pt}|}{} & \multicolumn{2}{c|}{Jigsaw} \\
\cline{2-3}
\multicolumn{1}{>{\centering}m{25pt}|}{} & \multicolumn{1}{m{17pt}|}{\textit{male}} & \multicolumn{1}{>{\centering}m{20pt}|}{\textit{female}} \\
\cline{1-3}
Baseline & 0.427 & 0.401  \\
Constrained   & \textbf{0.453} & \textbf{0.470} \\
\cline{1-3}
\end{tabular*}
\quad
\begin{tabular*}{9.9pc}{p{39pt}|p{16pt}|p{16pt}|p{16pt}|p{16pt}|p{16pt}|}
\cline{2-6}
\multicolumn{1}{>{\centering}m{15pt}|}{} & \multicolumn{5}{c|}{Twitter} \\
\cline{2-6}
\multicolumn{1}{c|}{} & \multicolumn{1}{c|}{\textit{Eng}} & \multicolumn{1}{c|}{\textit{Ita}} & \multicolumn{1}{c|}{\textit{Pol}} & \multicolumn{1}{c|}{\textit{Port}} & \multicolumn{1}{c|}{\textit{Spa}}\\
\cline{2-6}
& \textbf{0.711} & 0.315  & 0.263 & 0.230 & 0.329 \\
& 0.710 & \textbf{0.332}  & \textbf{0.300} & \textbf{0.335} & \textbf{0.335} \\
\cline{2-6}
\end{tabular*}
\quad
\begin{tabular*}{6.7pc}{>{\centering}m{55pt}|c|c|}
\cline{2-3}
\multicolumn{1}{>{\centering}m{55pt}|}{} & \multicolumn{2}{c|}{Wiki} \\
\cline{2-3}
\multicolumn{1}{>{\centering}m{55pt}|}{} & \multicolumn{1}{>{\centering}m{26pt}|}{\textit{'01-'14}} & \multicolumn{1}{>{\centering}m{26pt}|}{\textit{'15-'16}} \\
\cline{2-3}
& 0.703 & 0.710  \\
& \textbf{0.807} & \textbf{0.807} \\
\cline{2-3}
\end{tabular*}
\quad
\begin{tabular*}{10pc}{p{44pt}|p{16pt}|p{16pt}|p{16pt}|}
\cline{2-4}
\multicolumn{1}{>{\centering}m{25pt}|}{} & \multicolumn{3}{c|}{Gab} \\
\cline{2-4}
\multicolumn{1}{>{\centering}m{25pt}|}{} & \multicolumn{1}{m{15pt}|}{\textit{rel.}} & \multicolumn{1}{m{15pt}|}{\textit{race}} & \multicolumn{1}{m{15pt}|}{\textit{nat.}}\\
\cline{2-4}
& 0.240 & 0.198 & \textbf{0.473} \\
& \textbf{0.279} & \textbf{0.225} & 0.412 \\
\cline{2-4}
\end{tabular*}
\label{table2}
\end{table*}

\begin{table*}
\caption{Contingency tables of baseline (unconstrained) and fairness-constrained cyberbullying detection models calculated from the test sets of each experiment.}
\hspace{0.1cm}
\small
\begin{tabular*}{6.7pc}{|>{\centering}m{55pt}|c c|}
\cline{2-3}
\multicolumn{1}{>{\centering}m{55pt}|}{} & \multicolumn{2}{c|}{Jigsaw} \\
\cline{2-3}
\multicolumn{1}{>{\centering}m{55pt}|}{} & \multicolumn{1}{>{\centering}m{30pt}|}{Baseline model correct} & \multicolumn{1}{>{\centering}m{30pt}|}{Baseline model wrong} \\
\cline{1-3}
Constrained model correct & 48,802 & 5,849  \\
\cline{1-1}
Constrained model wrong   & 1,888   & 4,055 \\
\cline{1-3}
\multicolumn{3}{l}{$p < 0.001$ for all McNemar's tests.}
\end{tabular*}
\quad
\begin{tabular*}{6.7pc}{>{\centering}m{55pt}|c c|}
\cline{2-3}
\multicolumn{1}{>{\centering}m{55pt}|}{} & \multicolumn{2}{c|}{Twitter} \\
\cline{2-3}
\multicolumn{1}{>{\centering}m{55pt}|}{} & \multicolumn{1}{>{\centering}m{30pt}|}{Baseline model correct} & \multicolumn{1}{>{\centering}m{30pt}|}{Baseline model wrong} \\
\cline{2-3}
 \hfill \\ & 13,398 & 758  \\
 \hfill \\ & 539   & 1,422 \\
\cline{2-3}
\multicolumn{3}{l}{}
\end{tabular*}
\quad
\begin{tabular*}{6.7pc}{>{\centering}m{55pt}|c c|}
\cline{2-3}
\multicolumn{1}{>{\centering}m{55pt}|}{} & \multicolumn{2}{c|}{Wiki} \\
\cline{2-3}
\multicolumn{1}{>{\centering}m{55pt}|}{} & \multicolumn{1}{>{\centering}m{30pt}|}{Baseline model correct} & \multicolumn{1}{>{\centering}m{30pt}|}{Baseline model wrong} \\
\cline{2-3}
 \hfill \\ & 10,205 & 768  \\
 \hfill \\ & 299   & 424 \\
\cline{2-3}
\multicolumn{3}{l}{}
\end{tabular*}
\quad
\begin{tabular*}{1pc}{>{\centering}m{55pt}|c c|}
\cline{2-3}
\multicolumn{1}{>{\centering}m{55pt}|}{} & \multicolumn{2}{c|}{Gab} \\
\cline{2-3}
\multicolumn{1}{>{\centering}m{55pt}|}{} & \multicolumn{1}{>{\centering}m{30pt}|}{Baseline model correct} & \multicolumn{1}{>{\centering}m{30pt}|}{Baseline model wrong} \\
\cline{2-3}
 \hfill \\ & 3,582 & 222  \\
 \hfill \\ & 57   & 289 \\
\cline{2-3}
\multicolumn{3}{l}{}
\end{tabular*}
\label{table3}
\end{table*}

Results of each experiment on the test set including $F_1$ score, Matthews correlation coefficient, precision, recall, false negative equality difference, false positive equality difference, and total bias for baseline and fairness-constrained models can be examined in \textbf{Table 1}. In addition, percentage decrease in total bias (baseline model vs. constrained model) is also reported in the same table. Group specific FNRs and FPRs as well as corresponding biases of each group for all experiments are displayed in \textbf{Figure 2} for detailed inspection. \textbf{Table 2} reports Matthews correlation coefficients of classifiers for each group of interest. Finally, we report the contingency tables of unconstrained and constrained models, calculated over the same test sets of all experiments in \textbf{Table 3}. All McNemar's tests calculated from these contingency tables resulted in statistical significance with $p<0.001$, therefore the null hypothesis (``predictive performance of two models are equal'') can be rejected.

Our results (see Table 1) show that training neural network models with fairness constraints decrease the overall unintended bias of cyberbullying classifiers significantly, while increasing overall CB detection performance in terms of MCC. Total bias decrease with respect to unconstrained models corresponds to 78.7\%, 11.3\%, 94.9\%, and 27.5\% for Jigsaw, Twitter, Wiki, and Gab experiments, respectively. Total bias is mainly coming from the FNRs for Twitter and Wiki experiments while the main contributor of bias is the FPRs for Jigsaw and Gab experiments. Our approach manages to mitigate the total unintended bias for both situations. Models trained on the Twitter dataset (five languages as groups of interest) carry the highest amount of total bias. Figure 2 shows that high FPR of \textit{Spanish} and high FNR of \textit{Portuguese} and \textit{Polish} tweets are the main contributors to the bias. Models trained on the Gab dataset carry the second highest total bias especially due to high FPR of \textit{race} and \textit{religion} identity groups. Baseline model trained on the Wiki dataset exhibits very little bias among comments posted in the recent years vs. older comments. Guiding the model training with fairness constraints reduces that bias almost to zero.

Our results in Table 2 show that group-specific CB detection performance increases with fairness constraints for almost all groups as well. Furthermore, Table 3 shows that constrained models perform statistically significantly better than unconstrained models in terms of overall cyberbullying detection accuracy. For instance for the same Jigsaw test set, 1,888 observations have been correctly classified by the unconstrained model while being misclassified by the constrained model; however, 5,849 observations have been misclassified by the unconstrained model while being correctly classified by the constrained model. We believe future studies would benefit from similar contingency table analyses as well, especially if performance metrics of compared models are close to each other.

\section{DISCUSSION}

We show that we can mitigate the unintended bias of cyberbullying detection models by guiding the model training with fairness constraints. Our experiments demonstrate the validity of our approach for different contexts such as gender bias and language bias. Furthermore, we show that bias mitigation does not impair model quality, on the contrary, improves it. This improvement may be attributed to constraints serving as a regularization mechanism during training, similar to the common practice of constraining magnitudes of gradients or model weights. As our approach does not require modification of model architecture and does not need group labels during inference, we can conclude that the model is able to jointly learn the statistical properties of cyberbullying and group membership.

High bias reduction of 78.7\% and 94.9\% is achieved in cases where there are only two groups of interest, i.e. \textit{male} and \textit{female} for Jigsaw and \textit{old} and \textit{recent} for Wiki experiment. We suspect that this is because learning to classify cyberbullying incidents are easier when only a few groups need to satisfy the upper-/lower-bound constraints. The lowest bias reduction, 11.3\%, is achieved in the language bias mitigation experiment with the Twitter dataset where both unconstrained and constrained models perform considerably better for English tweets. Even though we utilize a multilingual language model, this is not unexpected as English is the most prevalent language in the training data of pre-trained models as well as in the Twitter dataset itself. Models also show better performance in detecting hate speech on the basis of nationality vs. race and religion due to high false positive rates among race and religion groups in the Gab experiment.

One interesting observation is that fairness constraints tend to lower overall FPR by sacrificing FNR. Consequently, precisions increase (as FNR is 1 - recall). This behaviour is meaningful in cases where the bias comes predominantly from FPED (e.g. Jigsaw and Gab), because there will be less room to deviate when FPRs are approaching to zero. However, we observe this phenomenon in Twitter and Wiki experiments as well and do not have an explanation to it. We also observe that constraints do not have to be satisfied perfectly for successful bias mitigation. For instance, Twitter and Gab experiments have several groups that do not fall into the allowed deviation region (green bands in Figure 2), yet still carry considerably less bias compared to unconstrained models.

Our bias mitigation approach has several advantages, especially in terms of generalization ability. First, it is agnostic to neural network model architecture, regardless of complexity. Second, it is agnostic to data type and number of identity groups. While previously proposed mitigation methods rely heavily on textual data and statistical balancing of it, our method can work with any modes of data including images, text, or network attributes. Another practical advantage of constrained training of neural network models is that it enables defining arbitrary performance constraints in terms of various rates including FNR, FPR, precision, recall, or accuracy. Certain real-life CB detection applications may not afford to miss any cyberbullying incidents while others may require minimum false-alarms. The former will benefit from an allowed lower bound constraint on the overall recall (i.e. recall should not decrease below the given value), while the latter application can employ an allowed upper bound constraint on the FPR. Our framework provides an uncomplicated mechanism to incorporate such constraints into training. Finally, as we address the problem of bias mitigation by adjusting the machine learning model rather than the data, our approach can be seamlessly combined with previously proposed approaches for further reduction of unintended biases.

Considering the high prevalence of cyberbullying in private conversations instead of public ones, our study shares the same limitations with the previous studies of automatic cyberbullying detection: data scarcity and data representativeness. There is a scarcity of annotated datasets that are available to public for research due to protection of users’ privacy and immense need of resources for quality annotations. Such scarcity is amplified considering the research question of this study, as datasets having annotations of both identity attributes and cyberbullying are further limited. Therefore, we acknowledge that statistical distributions may differ largely among research datasets as well as between research datasets and real-life phenomena. Models trained and validated on specific cohorts, including our models, may not generalize to other cohorts or real-life scenarios. As discussed in \cite{rosa2019automatic}, this may be due to varying definitions of cyberbullying or varying methodologies for data collection and annotation. Furthermore, models trained on data collected from a certain period of time may not generalize to future scenarios due to distribution shift. Another limitation of our study is that we have adopted a broader definition of cyberbullying and treated each comment or post as an observation; ignoring the repetitiveness criteria of defining cyberbullying \cite{rosa2019automatic}.

Future work includes investigating the performance of our bias mitigation approach among sub-groups (e.g. young Hispanic males) and combining our approach with previous studies that counteract bias by strategically altering the training data. Validating our approach in a cross-dataset manner to test whether our mitigation approach generalizes across different cohorts is also in the scope of future work. As we have not performed any hyper-parameter tuning in this study, we also expect detection and bias mitigation performance improvements by investigating the parameter choices. For instance, we have neither preprocessed the textual data, nor investigated different network architectures. We also chose the same amount of allowed deviation from overall FNR/FPR as a fairness constraint for each group as we aimed to decrease the total bias in our experiments. Setting different thresholds for each group may lead to even more equitable models (yet increases the number of hyper-parameters to tune).

Considering the rate of utilization of algorithms to all aspects of our lives, the demand from individuals for ethical and fair algorithmic decision making is expected to increase. Failing to mitigate unintended biases in such systems might negatively affect the business models of companies that utilize these technologies as well. Incorrect and unfair algorithmic content mediation (takedowns, filtering, and automatic content moderation) may infringe on users’ free speech. Implementing fair machine learning systems and publishing the outcomes could instill greater confidence in users by increasing transparency and trust. Therefore, we should elevate our awareness for developing more equitable algorithms and should continue to develop methods to reduce the existing biases at the same time.

% Mormons have had a complicated relationship with federal law. belief=True, toxic=False 
% "Yes for 2 years max. I am talking ages 2 to 14. And don't say breast feed for over 2 years - weird. :(". all false 
% This is one of the many reasons I don't buy into the bible...morons like this are trying to convince me they know what the ultimate deity is that created this entire universe?  Please... belief=True, toxic=True 
% WTF do singers know about anything beyond their craft? toxic=True, belief=False

\section{CONCLUSION}

In this study, we have aimed to mitigate the bias of cyberbullying detection models. For this purpose, we proposed a scheme to train cyberbullying detection models under fairness constraints. With our approach, we demonstrated that various types of unintended biases can be successfully mitigated without impairing the model quality. We believe our work contributes to the pursuit of more equitable machine learning solutions for cyberbullying detection. Future directions include combining our approach with other bias mitigation methods and establishing a comprehensive framework for unbiased machine learning algorithms in cyber-social health applications.

\bibliographystyle{IEEEtran}
\bibliography{references}

% \begin{IEEEbiography}{Oguzhan Gencoglu}{\,}is with the Faculty of Medicine and Health Technology, Tampere University, Finland. His research interests include machine learning and health informatics. Contact him at oguzhan.gencoglu@tuni.fi or at oguzhangencoglu90@gmail.com.
% \end{IEEEbiography}

\end{document}